\documentclass{article}
\usepackage{spconf,amsmath,graphicx}
\usepackage{stfloats}
\usepackage[mathscr]{euscript}
\usepackage{makecell}
\usepackage{verbatim}
\usepackage{color}
\usepackage{caption}
\usepackage{subcaption}
\usepackage{array}
\newcommand{\jy}[1]{\textcolor{black}{#1}}
\newcommand{\zl}[1]{\textcolor{black}{#1}}
\newcommand{\cao}[1]{\textcolor{black}{#1}}
\newcommand{\ye}[1]{\textcolor{black}{#1}}
\newcommand{\zll}[1]{\textcolor{red}{#1}}

\title{A2S-NAS: ASYMMETRIC SPECTRAL-SPATIAL NEURAL ARCHITECTURE SEARCH FOR HYPERSPECTRAL IMAGE CLASSIFICATION}
\name{Lin Zhan$^1$, Jiayuan Fan$^*{}^2$\thanks{$^*$Corresponding author.}\thanks{This work was supported in part by the National Natural Science Foundation of China under Grant 62101137 and Grant
62071127, and in part by the Zhejiang Lab Project under Grant 2021KH0AB05.}, Peng Ye$^1$, Jianjian Cao$^1$}
\address{$^1$School of Information Science and Technology, Fudan University, Shanghai, China\\
$^2$Academy for Engineering and Technology, Fudan University, Shanghai, China}
\begin{document}
%\ninept
%
\maketitle
\begin{abstract}
%\textcolor{red}{Hyperspectral image (HSI) plays an important role in the human exploration of the Earth.}
%Hyperspectral images (HSI) play a very important role in the human exploration of the Earth. 
%Recently, the research of HSI classification based on deep learning has achieved promising results. Influenced by the research on Neural Architecture Search, some automatically designed HSI classification networks have been proposed. 
%However, HSI data has unique physical characteristics that most existing neural networks are too fixed to handle. 
%Most existing deep learning based networks are too fixed to handle hyperspectral image (HSI) data due to its unique physical characteristics. On the one hand, the size and shape of different feature classes vary greatly from one another, so a single receptive field is difficult to extract \textcolor{red}{distinctive} features. On the other hand, \textcolor{red}{plenty of} methods do not pay attention to the asymmetric of HSI in the amount of information in the spectral and spatial dimensions. \textcolor{red}{To address this issue, we propose a hierarchical spectral-spatial neural architecture search (NAS) algorithm for HSI Classification, with an outer search space and an inner search space, by flexibly operating multiple receptive fields and asymmetric pooling on the spectral-space dimension.} The proposed method have been demonstrated on two challenging HSI benchmarks including Indian Pines and \jy{Houston} 2013, outperforming handcraft as well as NAS-based networks.
\jy{Existing deep learning-based hyperspectral image (HSI) classification works still suffer from the limitation of the fixed-sized receptive field, leading to difficulties in distinctive spectral-spatial features for ground objects with various sizes and arbitrary shapes. Meanwhile, plenty of previous works ignore asymmetric spectral-spatial dimensions in HSI. To address the above issues, we propose a multi-stage search architecture in order to overcome asymmetric spectral-spatial dimensions and capture significant features.
\zl{%The proposed method utilizes a multi-stage search architecture in order to overcome asymmetric spectral-spatial dimensions and capture significant features.
First, the asymmetric pooling on the spectral-spatial dimension maximally retains the essential features of HSI. Then, the 3D convolution with a selectable range of receptive fields overcomes the constraints of fixed-sized convolution kernels. Finally, we extend these two searchable operations to different layers of each stage to build the final architecture.} Extensive experiments are conducted on two challenging HSI benchmarks including Indian Pines and Houston University, and results demonstrate the effectiveness of the proposed method with superior performance compared with the related works.}

%Hyperspectral images (HSI) play a very important role in the human exploration of the Earth. Recently, the research of HSI classification based on deep learning has achieved promising results. Influenced by the research on Neural Architecture Search, some automatically designed HSI classification networks have been proposed. However, HSI data has unique physical characteristics that most existing neural networks are too fixed to handle. On the one hand, the size and shape of different feature classes vary greatly from one another, so a single receptive field is difficult to extract distinguishing features. On the other hand, many methods do not pay attention to the asymmetric of HSI in the amount of information in the spectral and spatial dimensions. To solve this issue, we propose a hierarchical spectral-spatial NAS algorithm for HSI Classification, including an outer search space and a inner search space. Multi-receptive field and asymmetric pooling on spectral-spatial dimension are provided in a flexible way. The proposed method have been demonstrated on two challenging HSI benchmarks including Indian Pines and HoustonU 2013, outperforming handcraft as well as NAS-based networks.
\end{abstract}

%
%\vspace{-8pt}
\section{Introduction}
\label{sec:intro}

%Hyperspectral image (HSI) classification, which categorizes each pixel into predefined classes by analyzing its spatial and spectral characteristics, is an frontier application of remote sensing. This valuable mission is the basis for practical applications such as vegetation research, fine agriculture, ocean exploration, and defense and security\cite{adam2010multispectral,teke2013short,lu2013determining}. HSI data, with the characteristic of space-spectral integration, contain hundreds of continuous narrow spectral channels which include a large amount of spectral information. The high-dimensional spectral data, however, poses a great challenge to HSI classification - the classification accuracy decreases as the dimensions of the input data increases, which is known as the Hughes phenomena~\cite{camps2006composite}.\\
%\captionsetup[figure]{labelfont=bf,textfont=normalfont,singlelinecheck=off,justification=raggedright}

Hyperspectral image classification is an advanced task within the field of remote sensing, which categorizes each pixel into predefined classes by analyzing its spectral and spatial characteristics. This valuable mission is the basis for practical applications such as vegetation research, fine agriculture, ocean exploration, and defense and security~\cite{teke2013short,lu2013determining}. HSI data has the characteristic of spectral-spatial integration, consisting of hundreds of continuous narrow spectral channels, which preserve a large amount of spectral information. However, the high-dimensional spectral data also poses a great challenge to the HSI classification task - the classification accuracy increases first as the dimensions of the input data increase. After reaching a certain extreme value, decreases as the dimensions of the input data increase, which is also known as the Hughes phenomena~\cite{hughes1968mean}.

To solve this issue, several traditional feature extraction methods based on principal component analysis ~(PCA)~\cite{prasad2008limitations}, independent component analysis~(ICA)~\cite{li2011locality}, and linear discriminant analysis~(LDA)~\cite{bandos2009classification} are proposed.
Additionally, support vector machines (SVM)~\cite{camps2005kernel, sun2014supervised}~are also a popular way to handle HSI classification. With the continuous development of deep learning, \ye{neural network based} methods such as convolutional neural networks (CNN) have attracted the interest of HSI researchers due to their superior performance in solving computer vision problems~\cite{li2014classification, pan2021unsupervised}.
Compared with the traditional methods, the deep learning-based method can automatically obtain high-level features of the HSI data through the training process without feature engineering for classification feature selection, which enables classification models to represent the features of datasets better and simultaneously improve classification accuracy. Nonetheless, the current deep learning-based HSI classification methods still have some unsolved challenges.

Most existing neural networks are too fixed to handle HSI classification problems. On the one hand, the ground objects in the HSI data with various sizes and arbitrary shapes require variable and flexible receptive fields. Also, asymmetric spectral-spatial dimensions in HSI need to be regarded seriously. On the other hand, the existing HSI datasets have different physical characteristics, such as spatial resolution, spectral range, and the number of bands. Expert-designed neural networks are not well-equipped to handle these differences. Neural architecture search (NAS) is a way to automate the design process of neural network architecture~\cite{zoph2016neural}, reducing the required effort and expert knowledge compared to designing the architecture artificially for each dataset. The mainstream search strategies of early NAS approaches are evolutionary algorithm~\cite{xie2017genetic} based and reinforcement learning~\cite{zoph2016neural} based approaches, which usually require hundreds of GPU days or large amount of computational resources to be consumed. Differentiable Architecture Search (DARTS)~\cite{liu2018darts} approach is proposed to relax the search space to a continuous domain so that it can be optimized by gradient descent, effectively reducing the complexity of NAS tasks. In recent years, some NAS-based HSI classification methods have been proposed~\cite{chen2019automatic,wang2022unified}, but the alternative operations of these methods are adapted from classical algorithms and are not specifically designed for HSI data characteristics.

\begin{figure*}[htbp]
    \centering
    \includegraphics[width=1\textwidth]{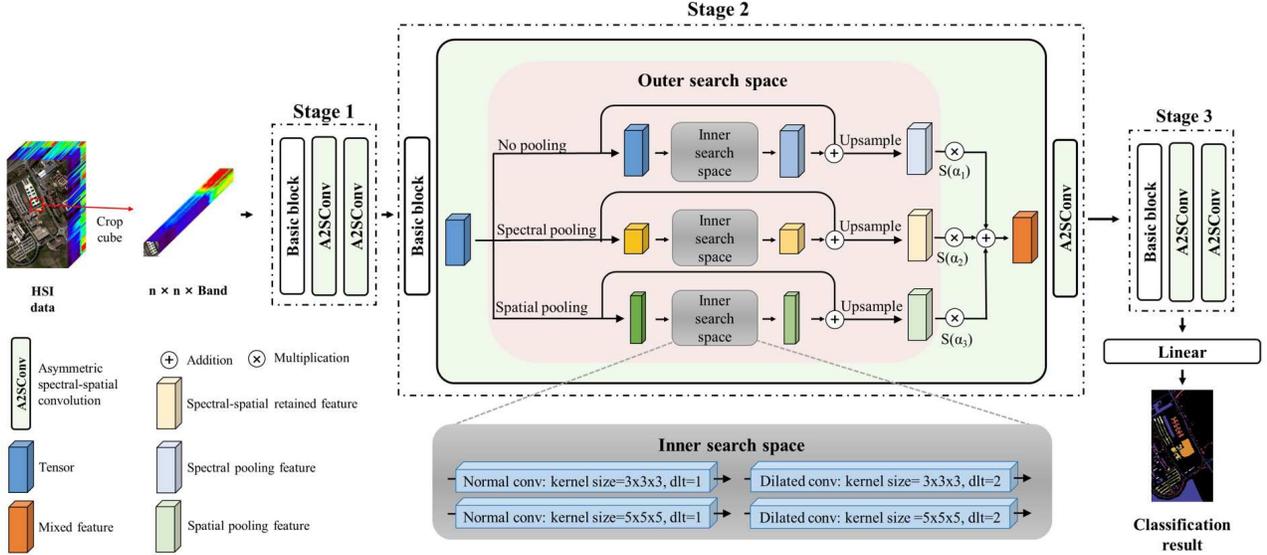}
    \vspace{-18pt}
    \captionsetup{font=small}
    %\caption{Flowchart of the proposed A2S-NAS. The main architecture consists of three stages, each of which contains two A2SConvs. \zl{The A2SConv, including an outer search space and an inner search space, search for the optimal settings for each layer, respectively. Outer search space is used to determine an asymmetric pooling operation, where three types of pooling are presented. Inner search space is adopted to decide the receptive field of 3D convolution.}} for each layer, respectively.
    \caption{Overview of the proposed Asymmetric Spectral-Spatial Neural Architecture Search (A2S-NAS). %The proposed method utilizes a multi-stage search architecture in order to overcome asymmetric spectral-spatial dimensions and capture significant features. 
    First, regular-sized cubes are cropped on the HSI dataset and fed into a multi-stage search architecture consisting of three stages. Second, Each stage contains a Basic block and two A2SConvs, where A2SConv is the proposed novel space to be searched. More specifically, the A2SConv includes an outer search space and an inner search space. Outer search space is used to determine an asymmetric pooling operation. Inner search space is adopted to decide the receptive field of 3D convolution. Finally, the output feature of the final stage is fed into a fully connected layer to obtain the classification result.}
    \label{fig:Fig. 1}
\end{figure*}

%To this end, we propose a novel \zl{asymmetric spectral-spatial NAS algorithm (A2S-NAS)} for HSI classification \textcolor{red}{task}. First, we propose a novel \zl{asymmetric spectral-spatial operator called A2SConv,} taking a more flexible way of providing appropriate receptive field and feature map resolution on spectral-spatial dimension. Second, \zl{a hierarchical spectral-spatial search structure} is dominated by the proposed A2SConv. Specifically, outer search space containing pooling on spectral and spatial dimensions and upsampling, searches for the appropriate feature map resolution. Inner search space is responsible for designing the perceptual field of the 3D convolution, respectively. In this work,the overall architecture of the network consists of three stages. 
\zl{To this end, we propose a novel asymmetric spectral-spatial NAS algorithm (A2S-NAS) for the HSI classification task. First, we propose a novel asymmetric spectral-spatial operator called A2SConv, including an outer search space and an inner search space. Second, a hierarchical spectral-spatial search structure is dominated by the proposed A2SConv. Specifically, outer search space contains searching for the appropriate asymmetric pooling operation and upsampling. The inner search space is responsible for designing the receptive field of the 3D convolution, respectively. Finally, we extend A2SConv to different layers of each stage in the overall network consisting of three stages.} The contributions of this paper are summarized as follows:
\begin{comment}
    \item[1)] We develop a flexible and hierarchical search which is composed of multiple HSConv blocks. The joint outer search sapce and inner search space allows simultaneous search of feature extract that contains different receptive fields and feature map resolutions for objects with various shapes and sizes.
    \item[2)] The HSNet is proposed considering the different physical resolution in the spectral and spatial dimension of HSI data. A gradient-based NAS algorithm is used to search the final network in order that the HSConv is able to perform asymmetric pooling in spectral and spatial dimensions.
    \item[3)] The proposed HSNet have been demonstrated on two challenging HSI benchmarks including Indian Pines and HoustonU 2013, outperforming handcraft as well as NAS-based networks.
\end{comment}
\begin{itemize}
    \item[1)] A2S-NAS is proposed considering the asymmetric physical resolution in spectral-spatial dimensions of HSI data. A gradient-based NAS algorithm is used to search the final network in order that \ye{different layers are} able to perform \ye{suitable} asymmetric pooling in spectral-spatial dimensions.
    \vspace{-4pt}
    \item[2)] We further develop a flexible and hierarchical search that is composed of certain A2SConv blocks. The joint outer search space and inner search space of A2SConv allows simultaneous search of \ye{asymmetric spectral-spatial feature pooling and} feature extraction that contains multiple receptive fields for objects with various shapes and sizes.
    \vspace{-4pt}
    \item[3)] The proposed A2S-NAS \ye{has been verified} on two challenging HSI benchmarks including Indian Pines and Houston University through extensive experiments, outperforming the related works with superior performance.
\vspace{-8pt}
\end{itemize}
\section{Method}
\label{sec:format}
%\vspace{-3pt}
%In this section, we first present the main architecture of the proposed HSNet \textcolor{red}{(Sec.\ref{sec:HSNet})}. Then, \textcolor{red}{in Sec.\ref{sec:HSConv},} we explain how the proposed HSConv dominates the construction of outer search space and inner search space. Finally, we explain how HSConv respectively searches the spectral pooling or spatial pooling as well as the perceptive field of 3D convolution \textcolor{red}{(Sec.\ref{sec:DSS})}.

%\subsection{A2S-NAS
%\raggedright}
%\label{sec:A2S-NAS}
The flowchart of the proposed A2S-NAS is shown in Fig.~1. First, we randomly cut regular-sized ($19\times19\times\text{Band}$, which is set based on the experimental results) cubes on the HSI dataset and send these cubes into a multi-stage search architecture, which consists of three stages. In detail, Stage 1 contains a Basicblock and two A2SConv blocks. Stage 2 and Stage 3 both contain a Downsample block and two A2SConv blocks where A2SConv is the designed novel space to be searched. 
%each \textcolor{red}{containing} two \textcolor{red}{A2SConv}.
%First, we randomly cut regular-sized patches on HSI dataset and send these patches into the main architecture\textcolor{red} consists of three stages, each of which contains two HSConvs.
Second, A2SConv modules that have an outer search space and an inner search space are used to conduct hierarchical spectral-spatial search.
%As mentioned earlier, the outer search space searches for an asymmetric pooling operation. The inner search space searches for a proper receptive field of 3D Convolution for the final network. 
Finally, the feature map produced by the last layer of the search part is fed into a pre-defined classifier composed of a linear layer to obtain the final classification result.
\vspace{-8pt}
\subsection{Asymmetric Spectral-Spatial Convolution
\raggedright}
\label{sec:A2SConv}
%As \textcolor{red}{illustrated} in Fig 1, the backbone of proposed network has three stages. 
As \cao{illustrated} in Fig.~1, the outer search space of A2SConv (pink zone in Fig.~1) searches for an asymmetric pooling operation. The inner search space of A2SConv (gray zone in  Fig.~1) searches for a proper receptive field of 3D convolution for the final \ye{HSI classification} network. \\
%The mainstream search strategies of early NAS approaches are evolutionary algorithm\cite{xie2017genetic} based and reinforcement learning\cite{zoph2016neural} based approaches, which usually require hundreds of GPU days or large amount of computational resources to be consumed. Therefore, researchers have worked hard to pursue how to reduce the computational effort and speed up the search. Differentiable Architecture Search (DARTS) approach was proposed in 2018\cite{liu2018darts}, which relaxes the search space to a continuous domain so that it can be optimized by gradient descent, effectively reducing the complexity of NAS tasks.
%In this paper, we use gradient-based search strategy with Beta-Decay regularization loss\cite{ye2022b}, which is a improving DARTS-based method without extra changes or cost. 
\textbf{Outer Search Settings.}
The outer search space includes asymmetric pooling and upsampling operation, aiming to reduce the high computational cost of processing high-dimensional HSI images while ensuring no loss of spectral and spatial features. In this work, we provide three different operations: no pooling, spectral pooling utilizing average pooling with stride (2,1,1), and spatial pooling utilizing average pooling with stride (1,2,2), respectively. \zll{}In contrast to traditional methods, we are searching for the corresponding pooling method \ye{(i.e., pooling dimension)} for each layer of the network, rather than simply pooling on a fixed dimension or the whole dimension, which is proved to play an important role in HSI classification network.\\
\textbf{Inner Search Settings.}
The physical characteristics of the HSI dataset introduced above inspire us that networks with multi-receptive fields can extract features of ground objects with various sizes and shapes. \ye{To this end, the proposed} inner search space consists of 3D convolutions with multi-receptive fields. In this work, we provide four different operations depending on the kernel size and dilatation rate, including 3D convolution with kernel size 3 or 5 and dilatation rate, which are chosen from 1 and 2, equal to normal convolution and dilated convolution. 
\vspace{-8pt}
\subsection{Multi-Stage Search Strategy
\raggedright}
\label{sec:DSS}
In contrast to the gradient-based 3D-ANAS~\cite{zhang20213}, the basic block of our search \ye{space} is the proposed \zl{A2SConv} itself, rather than a directed acyclic cell. In addition, we adopt a gradient-based search strategy with Beta-Decay regularization loss, which is an improving DARTS-based method without extra changes or cost~\cite{ye2022b}. The \zl{A2SConv} block for \ye{searching contains all candidate operations mentioned above, each of which has a corresponding learnable parameter $\alpha_{k}$. When} placing \ye{all} layers of the three stages together, the output of the i-th \zl{A2SConv} block to be searched is \ye{defined as}
\begin{equation}
\begin{aligned}
{\overline{O}(x)}^{(i)} = \sum_{k=1}^{|O|} F(\alpha_{k}^{(i)})O_{k}(x)\\
F(\alpha_{k}^{(i)}) = \frac{exp(\alpha_{k}^{(i)})}{\sum_{k^{’}=1}^{|O|} exp(\alpha_{k^{’}}^{(i)})}
\end{aligned}
\end{equation}
where $x$ is the input of the i-th \zl{A2SConv} block, $O$ is the candidate operation set, \ye{$F(\alpha_{k}^{(i)})$} denotes the softmax activation of \ye{the corresponding learnable parameter $\alpha_{k}^{(i)}$}~and~ ${\overline{O}(x)}^{(i)}$ is \ye{the mixed} output of the i-th \zl{A2SConv} block. 

\ye{Note that architecture parameters between different layers are not shared in our search strategy. Therefore, the task of differentiable architecture search is to learn a set of architecture parameters $\alpha$ for each A2SConv block} that makes the search process achieve optimal performance. During the search process, the architecture parameters $\alpha$ \ye{can be} optimized by the gradient descent algorithm. After the search, a compact network is obtained by keeping only the operators corresponding to the largest \ye{activated parameter} in each block. \ye{In this way, neural architecture search has evolved into the optimization process of a set of continuous variables.}

\ye{The architecture parameters $\alpha$ and network weights $\omega$} are jointly optimized following bi-level optimization objective on the training and validation sets according to Eq. (2),
\begin{equation}
\begin{aligned}
& \min_{\alpha}\mathscr{L}_{val}(\omega^*(\alpha),\alpha)+\lambda\mathscr{L}_{Beta}\\
& \text{ s.t. }\omega^*(\alpha) = \text{arg}\min_{\omega}\mathscr{L}_{train}(\omega,\alpha)\\
& \mathscr{L}_{Beta} = \text{log}(\sum_{k=1}^{|O|}e^{{\alpha}_{k}}) = \text{smoothmax}(\text\{\alpha\})
\end{aligned}
\end{equation}
\ye{where $\lambda = 1$ denotes the weight of beta decay regularization loss proposed in~\cite{ye2022b}}. Based on the performance of the validation set, the optimal network architecture parameters $\alpha^*$ is selected. Then the specific operator corresponding to each block is chosen according to the optimal parameters.

\section{Experiment}
\label{sec:majhead}

\subsection{Experimental Setup
\raggedright}
\label{ssec:subhead}
%\vspace{-4pt}
We use two challenging HSI benchmarks to evaluate the proposed A2S-NAS, including Indian Pines (IP) and Houston University (HU) . The IP dataset consists of 16 vegetation classes with $145 \times 145$ pixels and 200 hyperspectral bands. The HU dataset consists of 15 ground object classes with $349 \times × 1905$ pixels and 144 hyperspectral bands.

For the IP dataset and the HU dataset, we use 610 and 450 randomly selected samples for the architectural search, respectively. The SGD optimizer is used to optimize the network architecture parameters. The Adam optimizer and exponentially decaying learning rate strategy are used for the optimization of the network parameters.

To compare our searched model with other advanced models, we randomly select 50 samples in each category as the training set, 30 as the validation set, and the rest as the test set on the IP dataset. For the HU dataset, we randomly select 30 samples in each category as the training set, 30 as the validation set, and the rest as the test set.
%\vspace{-2pt}
\subsection{Experimental Results
\raggedright}
\label{ssec:subhead1}
%\vspace{-4pt}
We \ye{compare} the proposed A2S-NAS with the NAS-based approach 3DAutoCNN~\cite{chen2019automatic}, SSTN~\cite{zhong2021spectral} and 3D-ANAS~\cite{zhang20213}, which were published recently, as well as CNN based methods, such as 3DCNN~\cite{li2017spectral} and SSRN~\cite{zhong2017spectral}.

The classification results implemented on the IP dataset based on the six models are listed in Table 1. Fig. 2 shows the corresponding visualization results. The best results of each row are highlighted in bold.

\begin{figure}[htbp]
    \centering
    \begin{minipage}{.24\linewidth}
        \centering
        \includegraphics[width=2.0cm]{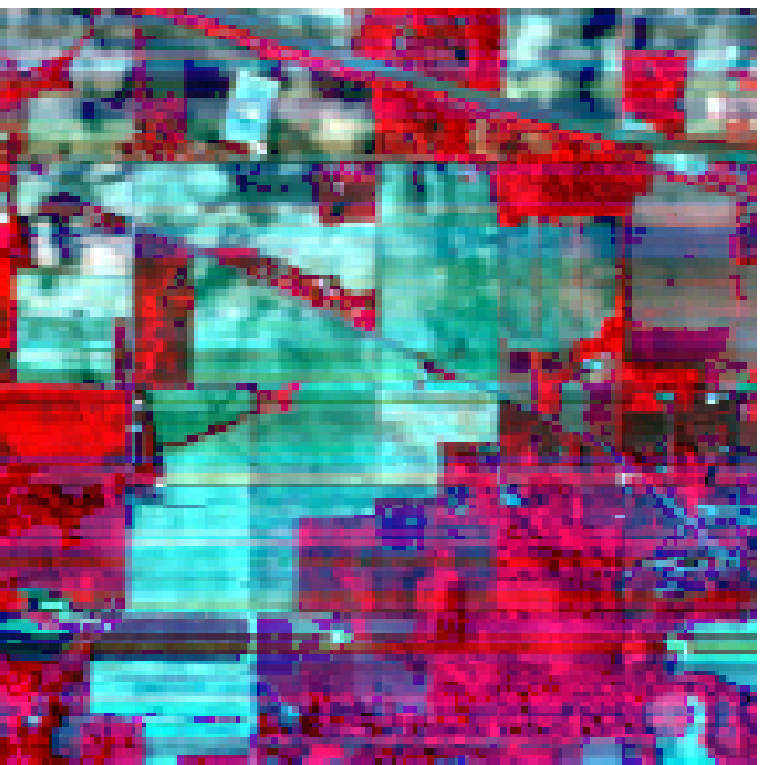}
        \centerline{(a)}\medskip
    \end{minipage}
    \begin{minipage}{.24\linewidth}
        \centering
        \includegraphics[width=2.0cm]{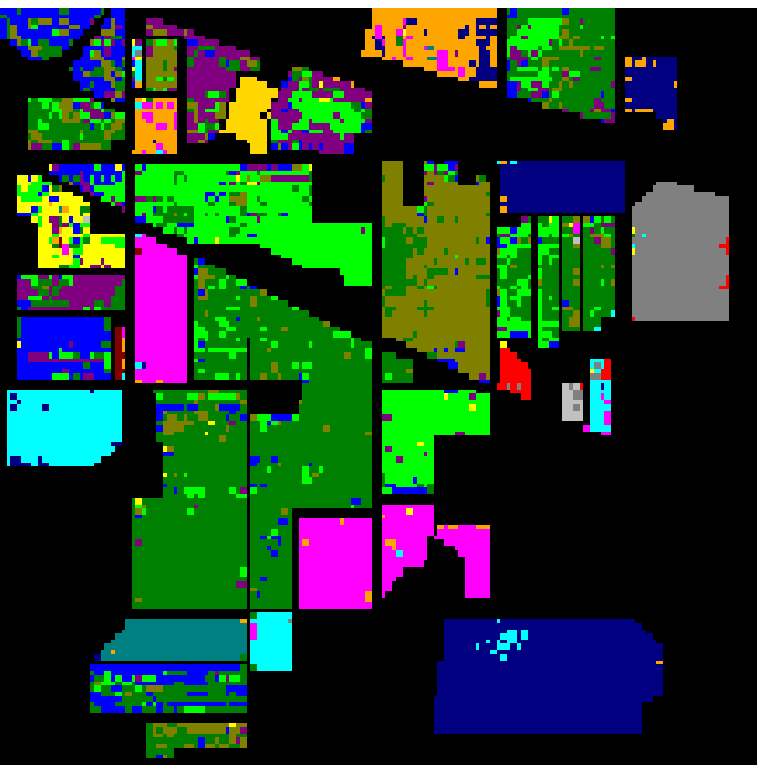}
        \centerline{(b)}\medskip
    \end{minipage}
    \begin{minipage}{.24\linewidth}
        \centering
        \includegraphics[width=2.0cm]{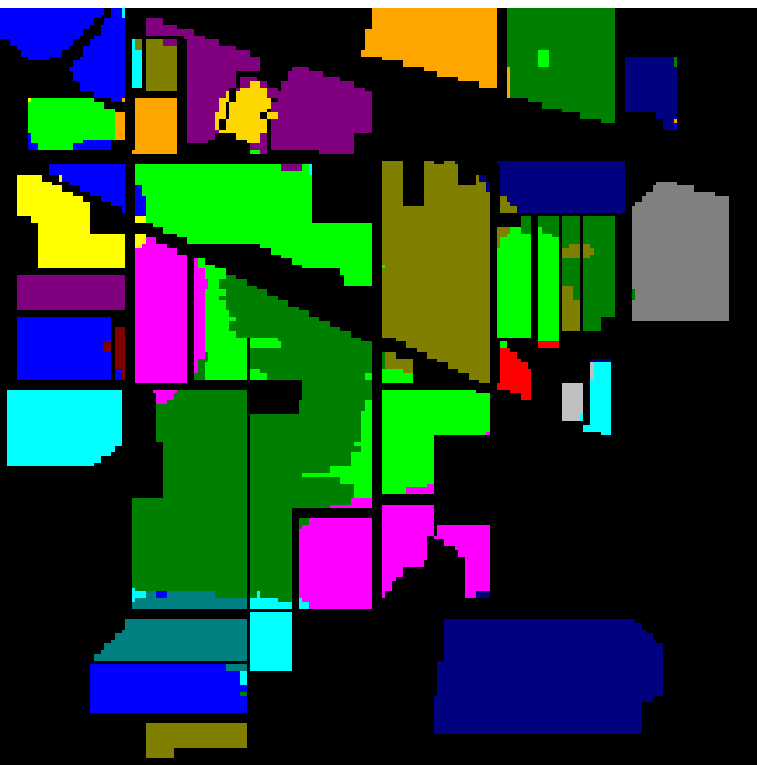}
        \centerline{(c)}\medskip
    \end{minipage}
    \begin{minipage}{.24\linewidth}
        \centering
        \includegraphics[width=2.0cm]{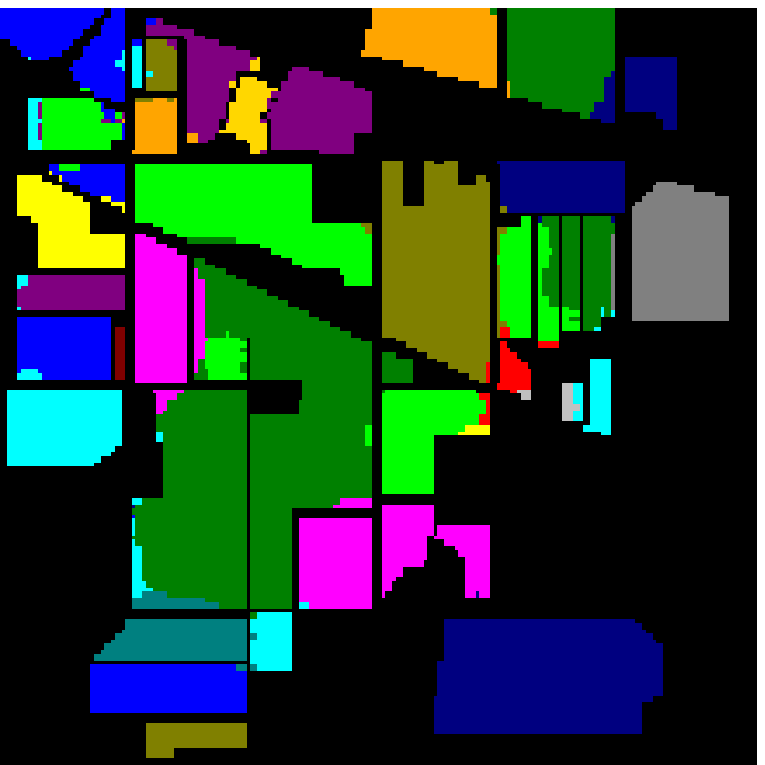}
        \centerline{(d)}\medskip
    \end{minipage}
    \begin{minipage}{.24\linewidth}
        \centering
        \includegraphics[width=2.0cm]{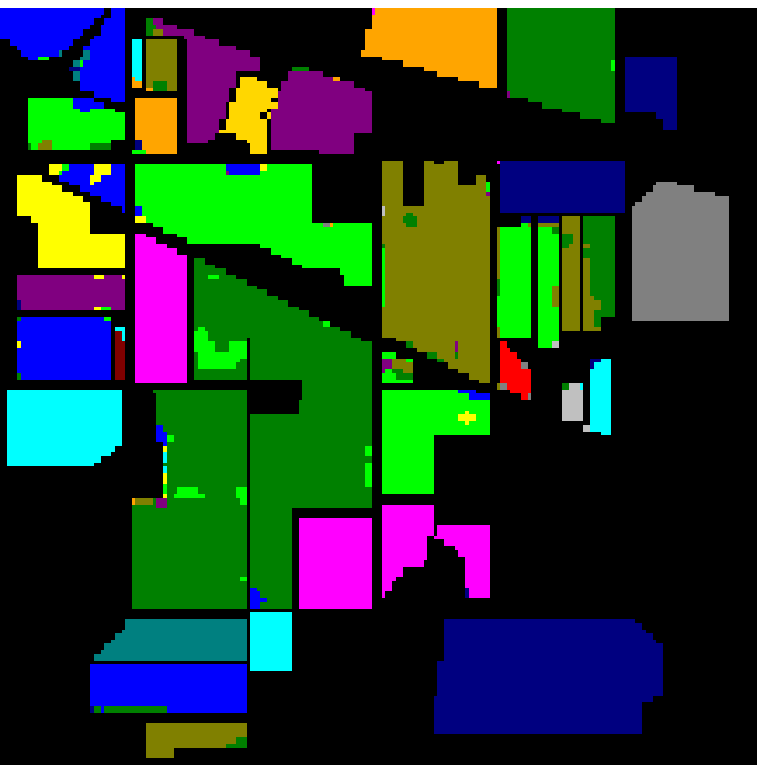}
        \centerline{(e)}\medskip
    \end{minipage}
    \begin{minipage}{.24\linewidth}
        \centering
        \includegraphics[width=2.0cm]{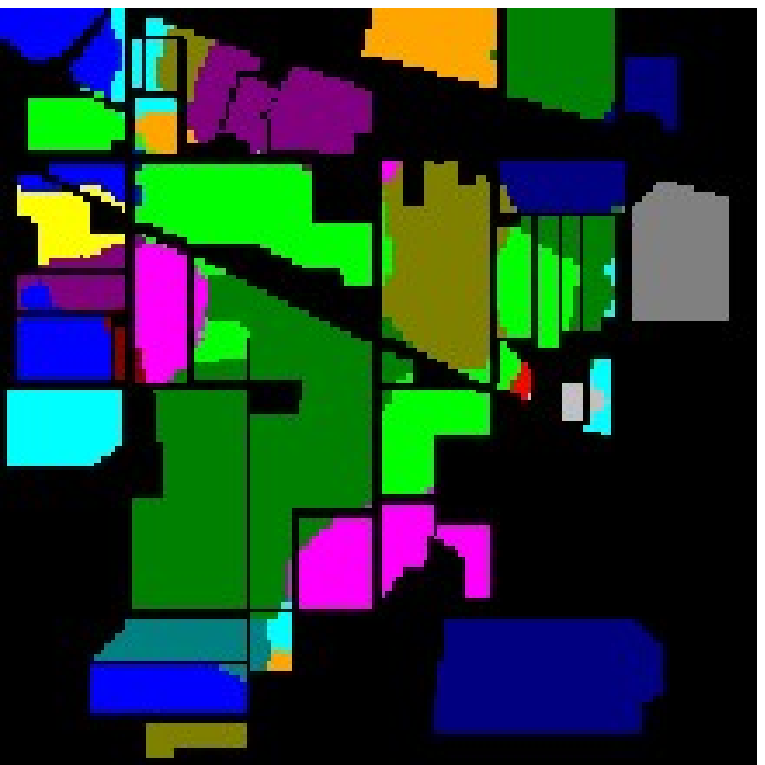}
        \centerline{(f)}\medskip
    \end{minipage}
    \begin{minipage}{.24\linewidth}
        \centering
        \includegraphics[width=2.0cm]{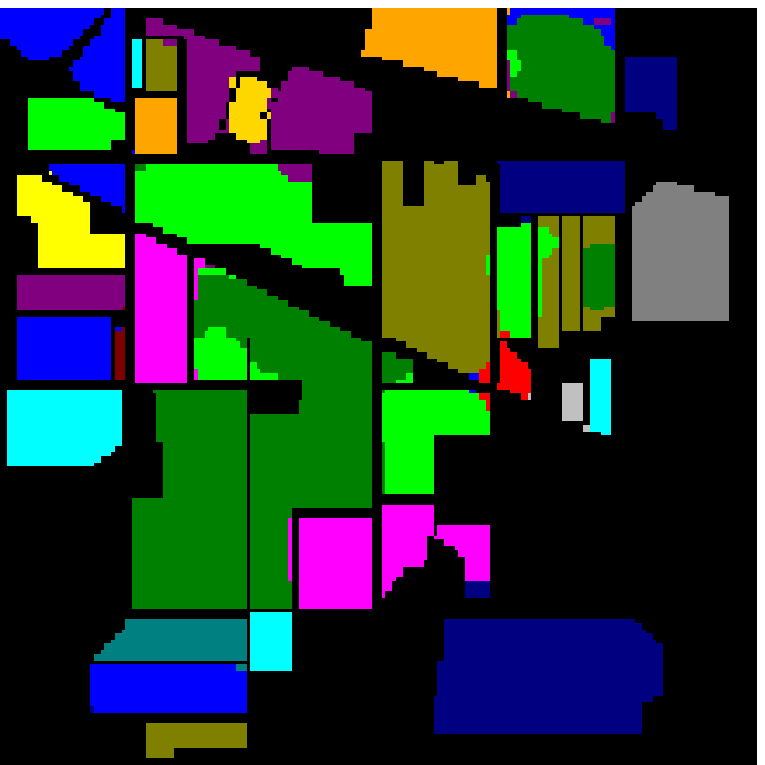}
        \centerline{(g)}\medskip
    \end{minipage}
    \begin{minipage}{.24\linewidth}
        \centering
        \includegraphics[width=2.0cm]{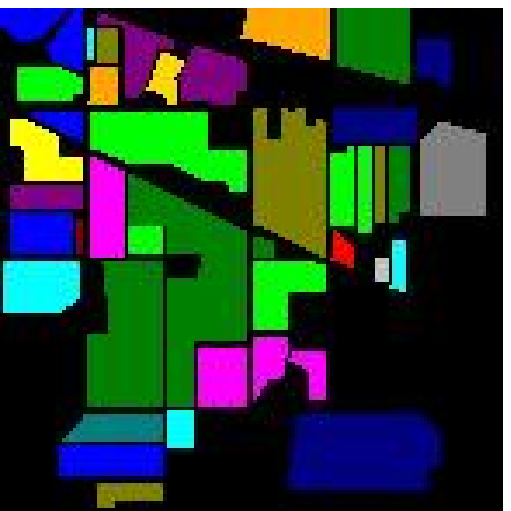}
        \centerline{(h)}\medskip
    \end{minipage}
\vspace{-6pt}	
\captionsetup{font=small}
\caption{Comparison of classification results of different models on IP dataset. (a)~False-color map. (b)~3DCNN, OA $= 76.83\%$. (c)~SSRN, OA $= 94.51\%$. (d)~3DAutoCNN, OA $= 94.09\%$. (e)~3D-ANAS, OA $= 95.91\%$. (f)~SSTN, OA $= 92.45\%$. (g)~A2S-NAS, OA $= 97.17\%$. (h)~Groundtruth.}
\label{fig2}
\end{figure}
\begin{figure}[t]
  \centering
  \centerline{\includegraphics[width=9.0cm]{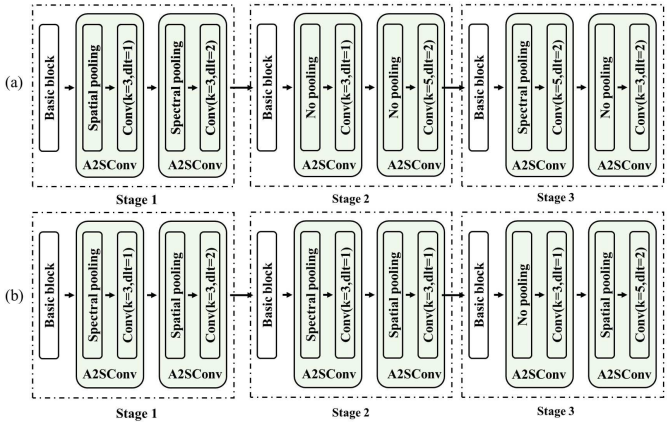}}
%  \vspace{1.5cm}
\vspace{-6pt}	
\captionsetup{font=small}
\caption{Final network for two HSI datasets. (a)~on IP dataset. (b)~on HU dataset.}
\vspace{-15pt}
\label{fig3}
\end{figure}
The classification results implemented on the HU dataset based on the six models are listed in Table 2. The best results of each row are highlighted in bold.

As can be seen in Tables 1 and 2, our proposed method A2S-NAS achieves the best classification results under all three evaluation metrics (OA for Overall Accuracy, AA for Average Accuracy, and Kappa for Kappa coefficient) among the six models on two HSI datasets. In particular, the classification results on class 11 Soybean-mintill and class 12 Soybean-clean of the IP dataset and class 4 Trees and class 9 Road of the HU dataset show a significant improvement of 3.55\%, 2.37\%, 2.61\% and 3.14\% \ye{, respectively}.

\begin{table}[h]\scriptsize
    \centering
    \captionsetup{font=small}
    \caption{Comparison of classification results of different models on IP dataset.}
    \vspace{-4pt}
    \begin{tabular}{c|c|c|c|c|c|c}
      \hline
      {Class} & {3DCNN} & {SSRN} & {\makecell[c]{3D\\Auto\\CNN}} & {\makecell[c]{3D-\\ANAS}} & {SSTN} & {\makecell[c]{A2S-\\NAS}}\\
      \hline
      OA(\%) & 71.65 & 94.51 & 94.09 & 95.91 & 92.45 & \textbf{97.17}\\
      AA(\%) & 83.28 & 93.01 & 96.66 & 95.35 & 93.68 & \textbf{97.05}\\
      Kappa & 67.70 & 93.66 & 93.28 & 95.26 & 91.29 & \textbf{96.73}\\
      \hline
      1 & 97.83 & \textbf{100.00} & 97.83 & 89.47 & 97.22 & 94.44\\
      2 & 50.00 & 84.26 & 89.92 & 90.96 & 80.12 & \textbf{92.17}\\
      3 & 53.15 & 97.95 & 95.66 & 93.84 & 95.48 & \textbf{98.22}\\
      4 & 89.78 & \textbf{100.00} & \textbf{100.00} & \textbf{100.00} & \textbf{100.00} & 97.81\\
      5 & 91.64 & 97.13 & 97.31 & \textbf{98.69} & 90.86 & 94.52\\
      6 & 93.81 & 99.52 & 98.90 & \textbf{99.68} & 93.49 & 98.89\\
      7 & 92.86 & 68.18 & \textbf{100.00} & 99.68 & \textbf{100.00} & \textbf{100.00}\\
      8 & 96.56 & \textbf{100.00} & \textbf{100.00} & 91.67 & \textbf{100.00} & \textbf{100.00}\\
      9 & 85.00 & 81.25 & \textbf{100.00} & \textbf{100.00} & 93.75 & 87.50\\
      10 & 70.64 & \textbf{97.82} & 91.67 & 81.25 & 89.45 & 94.27\\
      11 & 62.12 & 93.38 & 90.88 & 94.84 & 94.52 & \textbf{98.39}\\
      12 & 66.53 & 92.29 & 86.68 & 94.99 & 89.45 & \textbf{97.36}\\
      13 & \textbf{100.00} & \textbf{100.00} & \textbf{100.00} & 96.75 & \textbf{100.00} & \textbf{100.00}\\
      14 & 89.53 & 98.71 & 98.74 & \textbf{100.00} & 98.88 & \textbf{100.00}\\
      15 & 93.01 & 98.95 & 98.96 & \textbf{99.91} & 99.65 & 99.30\\
      16 & \textbf{100.00} & 78.67 & \textbf{100.00} & 98.25& 76.00 & \textbf{100.00}\\
      \hline
    \end{tabular}
\end{table}

\begin{table}[htbp]\scriptsize
\centering
\captionsetup{font=small}
    \caption{Comparison of classification results of different models on HU dataset.}
    \vspace{-4pt}
    \begin{tabular}{c|c|c|c|c|c|c}
      \hline
      {Class} & {3DCNN} & {SSRN} & {\makecell[c]{3D\\Auto\\CNN}} & {\makecell[c]{3D-\\ANAS}} & {SSTN} & {\makecell[c]{A2S-\\NAS}}\\
      \hline
      OA(\%) & 81.67 & 90.95 & 84.45 & 87.82 & 89.42 & \textbf{92.20}\\
      AA(\%) & 83.58 & 92.33 & 85.22 & 88.32 & 90.34 & \textbf{93.30}\\
      Kappa & 80.19 & 90.21 & 83.25 & 86.83 & 88.56 & \textbf{91.56}\\
      \hline
      1 & 95.43 & 91.78 & 87.50 & 87.22& 87.14 & \textbf{96.35}\\
      2 & \textbf{95.70} & 95.55 & 77.91 & 87.75 & 89.38 & 94.48\\
      3 & 97.55 & 96.60 & 92.74 & 84.26 & \textbf{98.91} & 96.10\\
      4 & 96.26 & 94.10 & 72.65 & 81.30 & 92.86 & \textbf{98.87}\\
      5 & 98.06 & 98.47 & 96.14 & 97.19 & \textbf{99.11} & 95.34\\
      6 & 92.11 & \textbf{98.21} & 84.86 & 88.81 & 90.68 & 97.49\\
      7 & 70.27 & \textbf{95.41} & 73.03 & 80.45 & 84.89 & 89.76\\
      8 & 70.87 & 77.51 & 76.64 & 79.32 & \textbf{84.31} & 68.93\\
      9 & 68.47 & 84.61 & 71.10 & 75.45 & 77.58 & \textbf{87.75}\\
      10 & 58.58 & 62.17 & \textbf{96.67} & 95.99 & 96.48 & 92.38\\
      11 & 78.35 & 96.75 & 92.31 & 95.77 & \textbf{99.47} & 96.75\\
      12 & 72.53 & \textbf{99.42} & 91.48 & 89.86 & 73.56 & 88.02\\
      13 & 68.88 & 94.41 & 85.80 & 94.08 & 83.39 & \textbf{95.63}\\
      14 & 95.58 & \textbf{100.00} & 90.65 & 99.25 & 99.78 & \textbf{100.00}\\
      15 & 94.99 & \textbf{100.00} & 88.85 & 95.71 & 97.56 & 99.59\\
      \hline
    \end{tabular}
\vspace{-6pt}
\end{table}
\vspace{-8pt}
\subsection{Architecture Analysis
\raggedright}
%\vspace{-2pt}
We analyze the search results on two datasets presented in Fig.~4, focusing on finding the effectiveness of hierarchical spectral-spatial search for asymmetric pooling methods in outer search space. First, different layers in the final network are able to perform 3D convolutions with different receptive fields, indicating that our method is flexible in terms of feature extraction. Second, the layers containing asymmetric pooling operations account for 50\% and 83.3\% of the two models, respectively. We believe that this consideration of asymmetric spectral-spatial feature pooling is promising.
\vspace{-3pt}
\section{Conclusion}
\label{conclutsion}
\vspace{-3pt}
Most deep learning-based HSI classification methods are too fixed to cope with the ground objects in the HSI data with various sizes and arbitrary shapes. In this paper, \zl{we propose the A2S-NAS for HSI classification}, taking a more flexible way of overcoming asymmetric spectral-spatial dimensions and capturing significant features.
%providing appropriate receptive field and feature map resolution on spectral-spatial dimension. 
%providing appropriate receptive field and feature map resolution on spectral-spatial dimension. 
Specifically, the proposed \zl{A2SConv} helps to construct a joint outer search space and inner search space, which was experimentally demonstrated to have superior classification ability. We believe this concern for asymmetric pooling operations and multi-receptive fields will have an impact on HSI classification research.
\vfill
\newpage
\bibliographystyle{IEEEbib}
\bibliography{strings,refs}

\end{document}